\documentclass[conference]{IEEEtran}
\IEEEoverridecommandlockouts
\usepackage{cite}
\usepackage{amsmath,amssymb,amsfonts}
\usepackage{algorithmic}
\usepackage{graphicx}
\usepackage{float}
\usepackage{textcomp}
\usepackage{xcolor}
\usepackage{amssymb}
\usepackage{hyperref}
\usepackage{subcaption}
\usepackage{dblfloatfix}
\usepackage{epstopdf}

\def\BibTeX{{\rm B\kern-.05em{\sc i\kern-.025em b}kern-.08em
    T\kern-.1667em\lower.7ex\hbox{E}\kern-.125emX}}

\begin{document}

\title{
Coupling of localization and depth data for mapping using Intel RealSense T265 and D435i cameras
}

\author{\IEEEauthorblockN{Evgeny Tsykunov}
\IEEEauthorblockA{
\textit{Skoltech}\\
Moscow, Russia \\
evgeny.tsykunov@skolkovotech.ru}
\and
\IEEEauthorblockN{Valery Ilin}
\IEEEauthorblockA{
\textit{Skoltech}\\
Moscow, Russia \\
valery.ilin@skoltech.ru}
\and
\IEEEauthorblockN{Stepan Perminov}
\IEEEauthorblockA{
\textit{Skoltech}\\
Moscow, Russia \\
Stepan.Perminov@skoltech.ru}
\and
\IEEEauthorblockN{Aleksey Fedoseev}
\IEEEauthorblockA{
\textit{Skoltech}\\
Moscow, Russia \\
aleksey.fedoseev@skoltech.ru}
\and
\IEEEauthorblockN{Elvira Zainulina}
\IEEEauthorblockA{
\textit{Skoltech}\\
Moscow, Russia \\
elvira.zainulina@skoltech.ru}}

\author{\IEEEauthorblockN{
Evgeny Tsykunov\IEEEauthorrefmark{1},
Valery Ilin\IEEEauthorrefmark{2},
Stepan Perminov\IEEEauthorrefmark{3},
Aleksey Fedoseev\IEEEauthorrefmark{4},
Elvira Zainulina\IEEEauthorrefmark{5}}
\IEEEauthorblockA{ 
Skolkovo Institute of Science and Technology, 
Moscow, Russia, 2020\\ \\
Emails: \IEEEauthorrefmark{1}evgeny.tsykunov@skolkovotech.ru,
\IEEEauthorrefmark{2}valery.ilin@skoltech.ru,
\IEEEauthorrefmark{3}stepan.perminov@skoltech.ru,\\
\IEEEauthorrefmark{4}aleksey.fedoseev@skoltech.ru,
\IEEEauthorrefmark{5}elvira.zainulina@skoltech.ru}}

\maketitle

\begin{abstract}
We propose to couple two types of Intel RealSense sensors (tracking T265 and depth D435i) in order to obtain a localization and 3D occupancy map of the indoor environment. 
We implemented a python-based observer pattern with multi threaded approach for camera data synchronization. 
We compared different point cloud (PC) alignment methods (using transformations obtained from tracking camera and from ICP family methods).
Tracking camera and PC alignment allows us to generate a set of transformations between frames. Based on these transformations we obtained different trajectories and provided their analysis.
Finally, having poses for all frames, we combined depth data. Firstly we obtained a joint PC representing the whole scene. Then we used Octomap representation to build a map.

\end{abstract}






\section{Introduction}


Our motivation is to navigate an autonomous robot in unknown environment. To perform the real-time localization and path planning, the robot first needs to obtain a map of it's environment. For this purpose we are proposing an algorithm that will fuse data from 2 different Intel RealSense cameras into 3D occupancy map, such as OctoMap \cite{hornung2013octomap}, providing similar to LIDAR solution at lower cost. We suggest to generate 3-dimensional rather than 2-dimensional map in order to make the solution more general and fit more number of applications (from aerial robotics to underwater robots). In order to fulfil that, we do not set any other constraints such as limit or allowed computational power.

In contrast to \cite{hornung2013octomap}, Guizilini ae al. in \cite{Guizilini_2018} proposed to create a continuous occupancy map using Hibert maps in real-time.
Another existing technology that generates height map for walking robot is presented in paper of Bayer et. al. \cite{bayer2019autonomous}. The height map is more simple to build than 3D occupancy map like OctoMap \cite{hornung2013octomap}, therefore it can be accomplished on a lightweight sing-board computer in real time. 



\section{Methodology}
Our main algorithm is provided bellow:
\begin{enumerate}
    \item Extract of the tracking (pose) data from T265 and depth data from D435i Intel RealSense cameras;
    \item Synchronize the data from cameras (by timestamps);
    \item Compare/build algorithms that is able to constantly update prior trajectory and map using the fusion of current data from depth and tracking cameras.
\end{enumerate}

\section{Hardware}
\begin{figure}[t]
\centering
\includegraphics[width=0.5\textwidth]{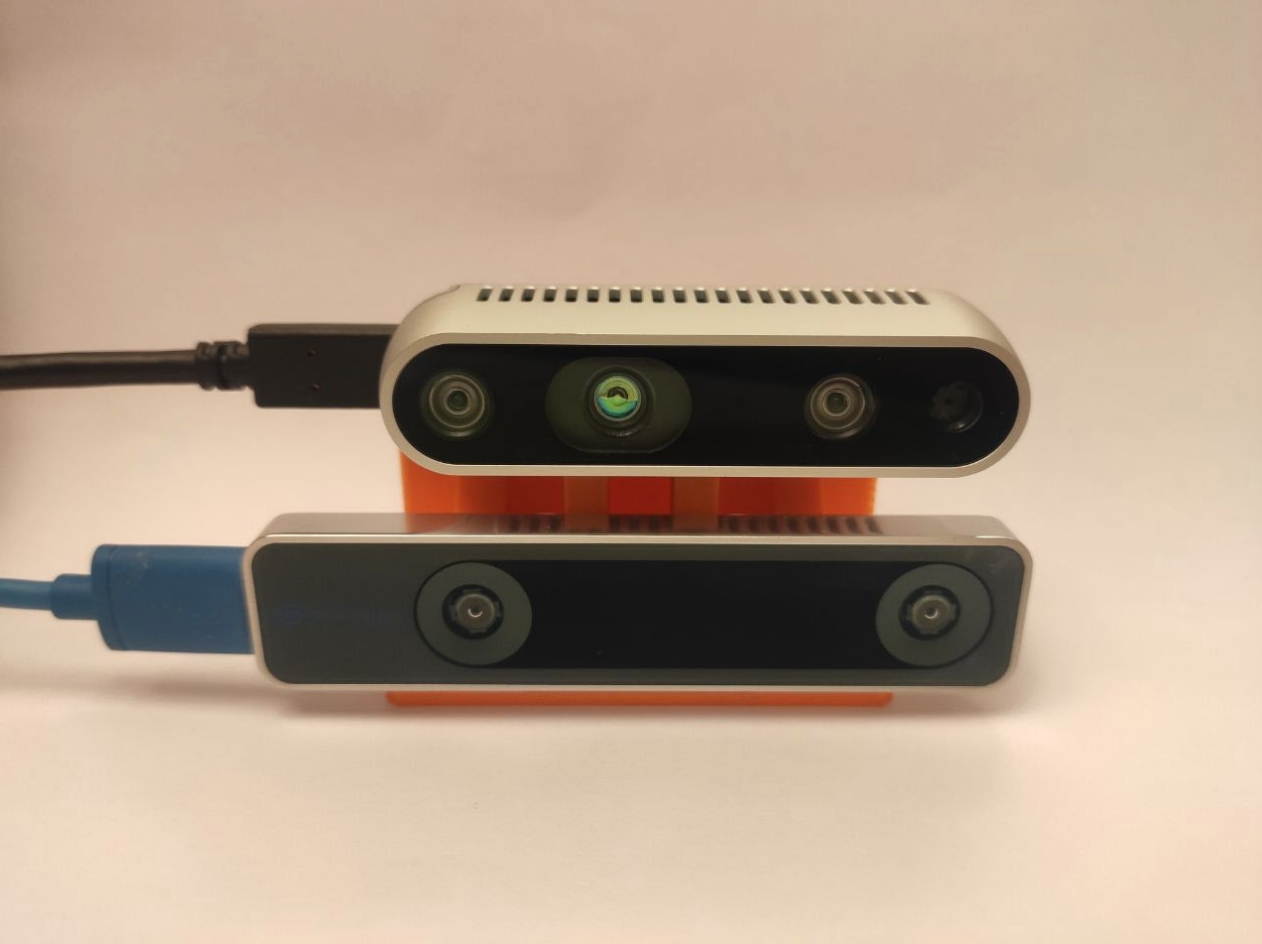}\label{true}
\caption{Sensor suite. Intel RealSense D435i RGBD (top). Intel Realsense T265 tracking camera (bottom) Designed mount provides known and constant transformation between cameras.}
\label{cameras}
\end{figure}

The sensor suite consists of two cameras. 
The first one (shown in Fig.\ref{cameras} in the bottom) is Intel Realsense T265 tracking camera which outputs the current pose (position and orientation) 200 time per second. The camera has two fisheye lenses with combined 163±5° FOV and BMI055 IMU Sensor on board. Visual Inertial Odometry from Intel is running on board.
The second (shown in Fig.\ref{cameras} on the top) is Intel RealSense D435i RGBD camera with IMU inside. The camera has global shutter and 3um x 3um pixel size. Active IR Stereo technology is used to obtain a depth data. Depth Field of View (FOV) $87^{\circ}±3^{\circ} x 58^{\circ}±1^{\circ} x 95^{\circ}±3^{\circ}$. Although we used only the depth channel with framerate 30.

\section{Collection of the data-set}
\begin{figure*}[t]
\centering
\includegraphics[width=\textwidth]{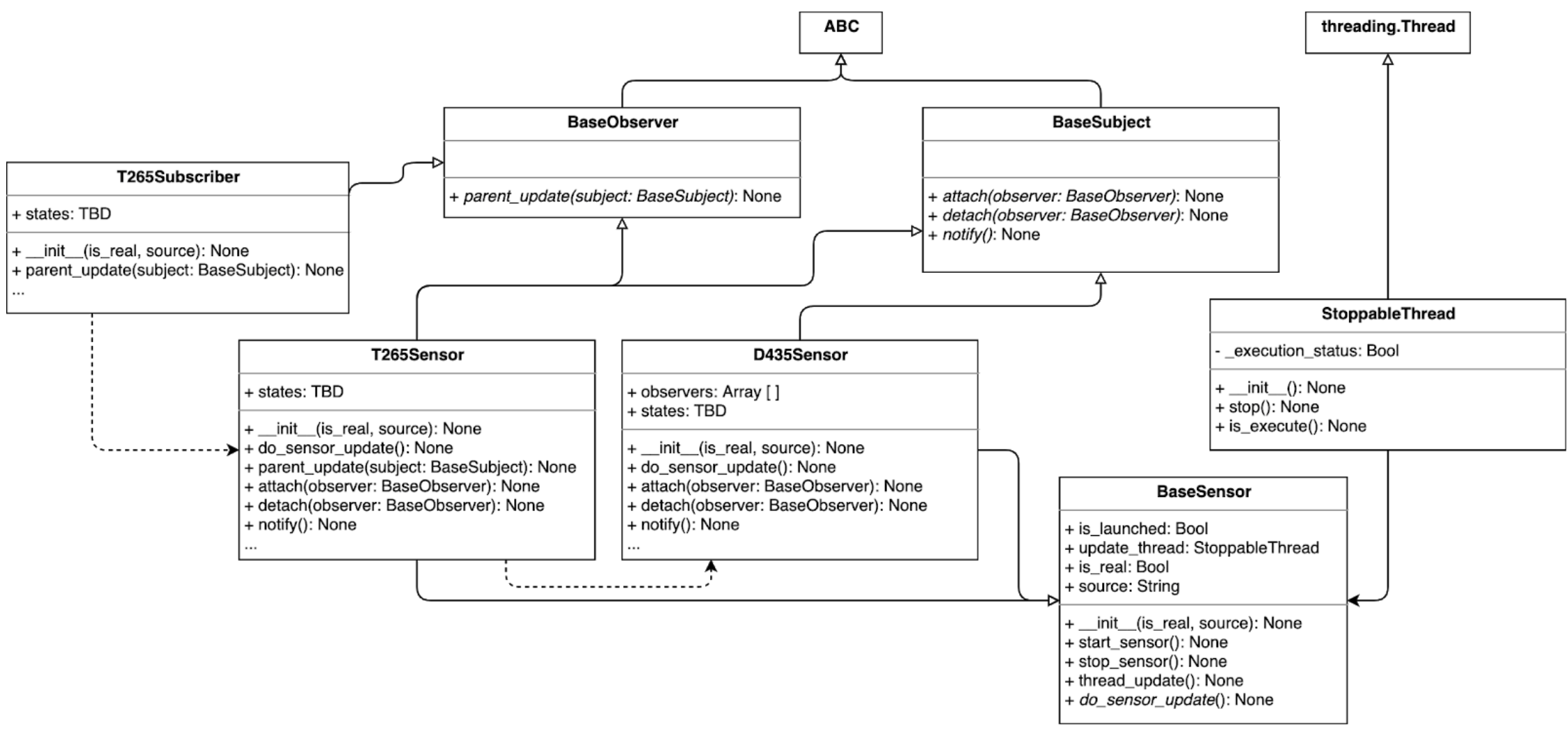}\label{true}
\caption{Python based observer programming pattern with multithreaded approach.}
\label{python_architechture}
\end{figure*}

In order to grab two streams (pose with 200 Hz and depth 30 Hz) in a synchronized way we implemented publisher-subscriber paradigm.
We built a Python application based on Observer programming pattern and multithreaded approach shown in Fig.\ref{python_architechture}. 

Code is presented by link\footnote{\url{https://github.com/IlinValery/perception_final_2020}}.
Whenever the slow depth stream receives a new frame, the fast pose stream is notified and save the closest pose.
In such way every depth frame has a pose assigned.
We collected an indoor dataset from both sensors and used it for the experiments.

\section{Used libraries}
\begin{enumerate}
    \item Open3D
    \item Pyrealsense
    \item Threading
    \item NumPy
    \item Matplotlib
    \item OpenCV - images visualisation
    \item PPTK - Point Cloud visualisation
    \item Octomap
    \item Pyglet, trimesh, glooey - Voxel Representation
\end{enumerate}

\section{Point cloud alignment}

\begin{figure}[H]
\centering
\includegraphics[width=0.4\textwidth]{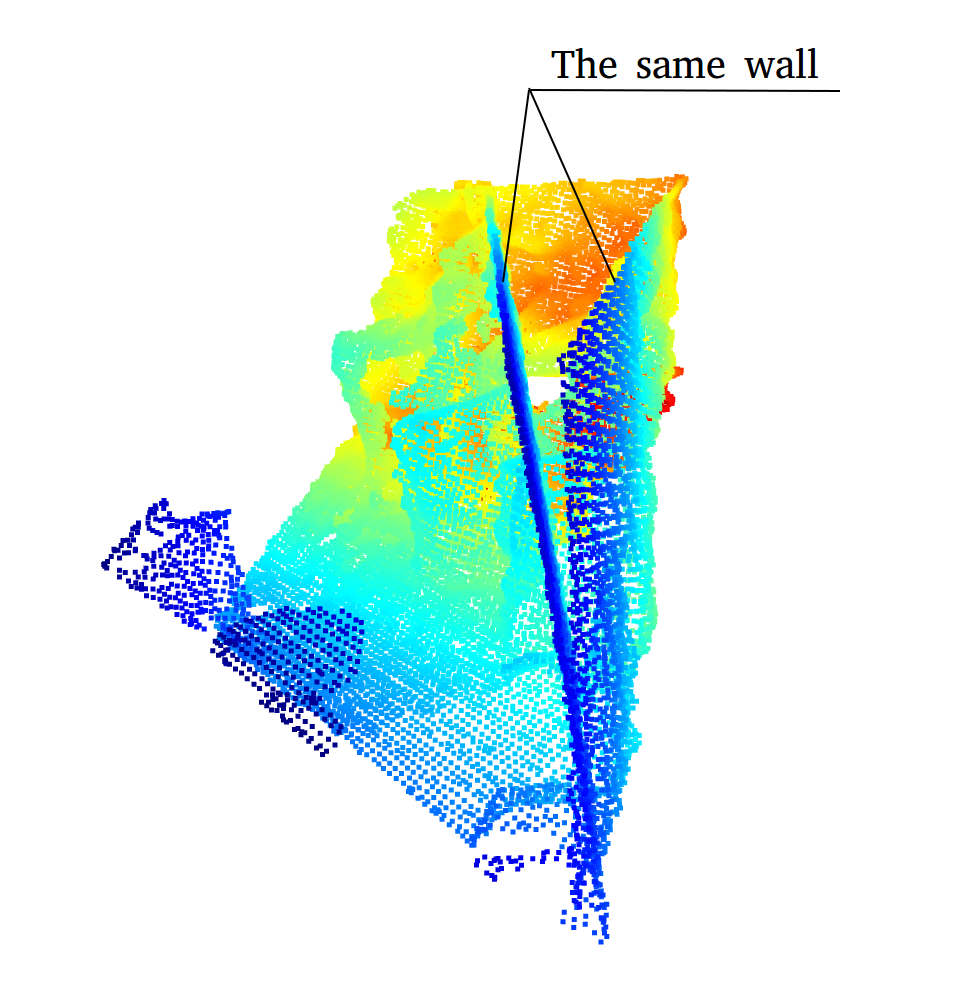}\label{true}
\caption{Misaligned PCs.}
\label{misaligned_pcs}
\end{figure}

From every depth frame, a point cloud was extracted with respect to camera local frame. Due camera motion all point clouds were getting misaligned in space with respect to the world frame (Fig.\ref{misaligned_pcs}). To eliminate the point cloud shifting the point cloud alignment was applied.

\begin{figure}[ht]
\begin{subfigure}{.5\textwidth}
  \centering
  \includegraphics[width=.8\linewidth]{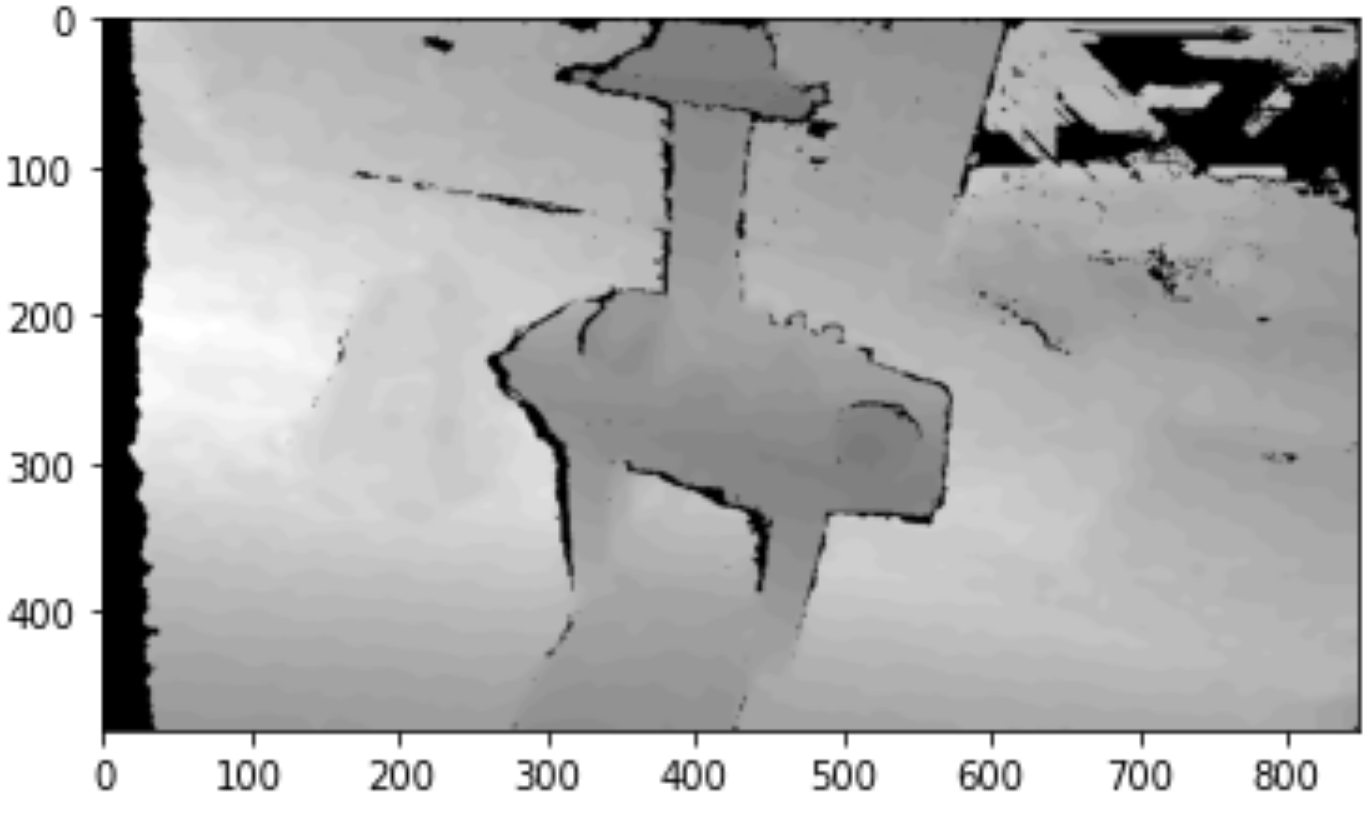}  
  \caption{Depth image from KeyFrame1}
  \label{fig:sub-first}
\end{subfigure}
\begin{subfigure}{.5\textwidth}
  \centering
  \includegraphics[width=.8\linewidth]{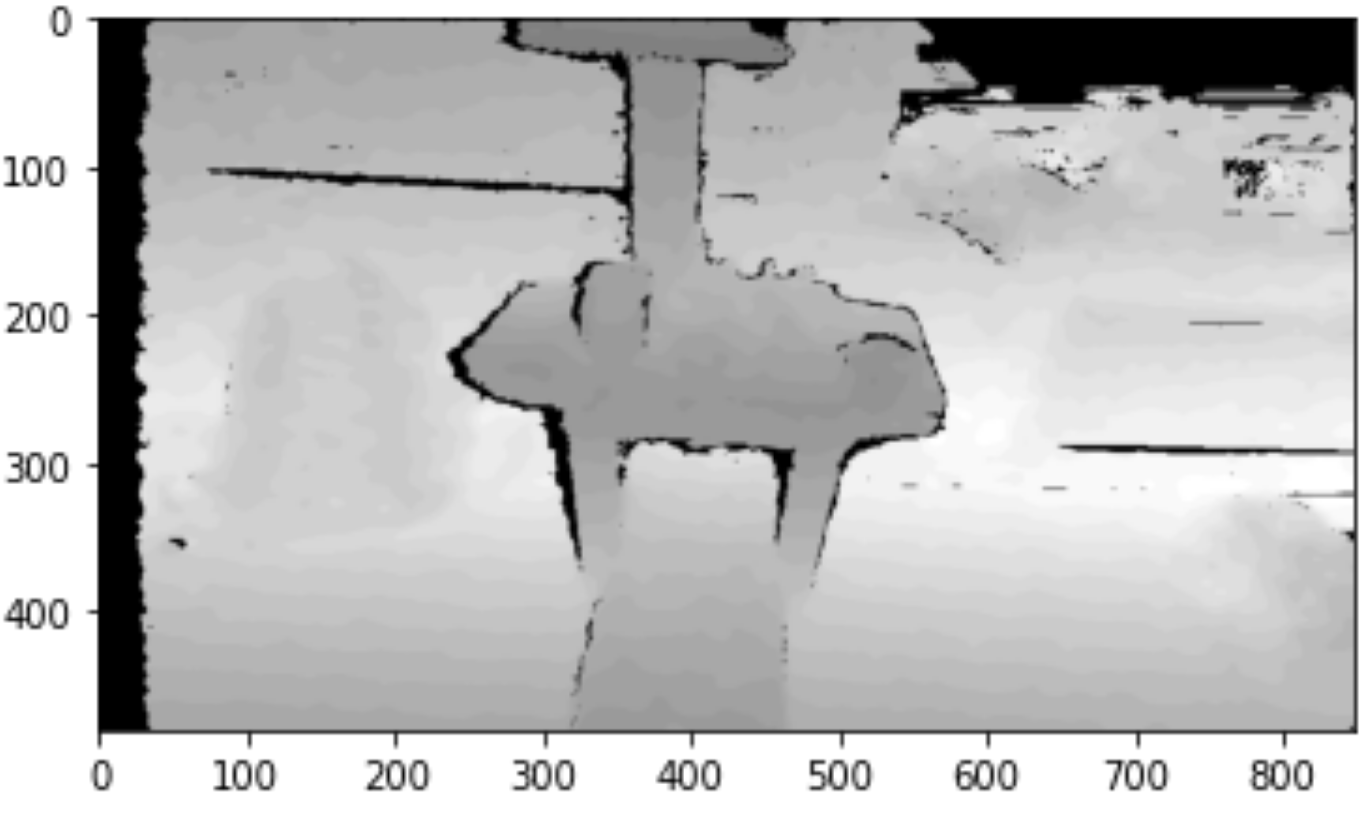}  
  \caption{Depth image from KeyFrame2}
  \label{fig:sub-second}
\end{subfigure}
\caption{Depth images from keyframes.}
\label{depth_frames}
\end{figure}

Two point clouds (from two keyframes, presented in Fig.\ref{depth_frames}) from the middle of our dataset with overlap of around 80\% (0.5 second in between) were selected for alignment. Two closest framesets from T265 and D435i (based on synchronization method) were extracted from two cameras.

\begin{figure}[h]
\centering
\includegraphics[width=0.35\textwidth]{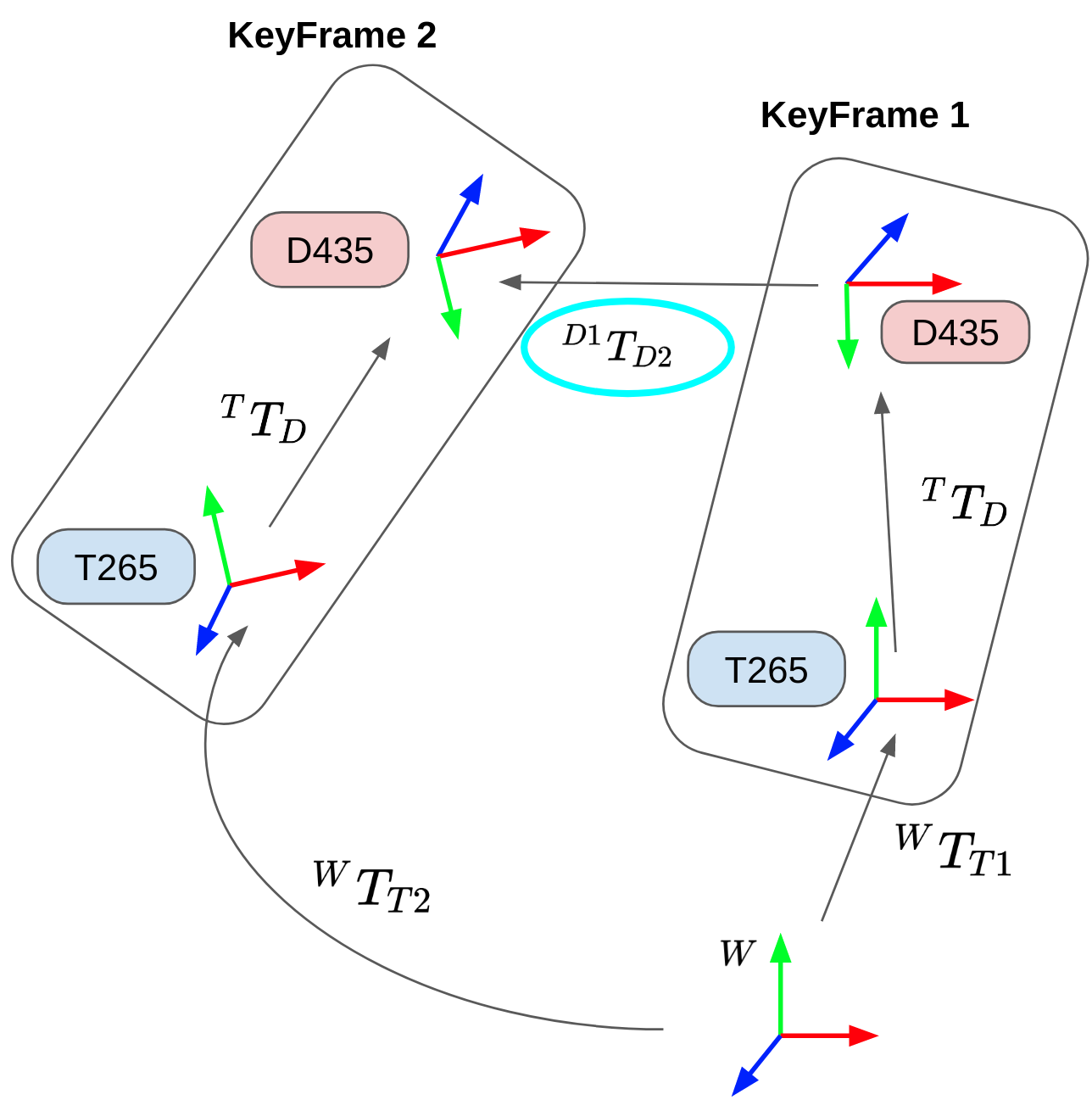}\label{true}
\caption{Transformations between frames.}
\label{transformations}
\end{figure}

First, we obtained transformation matrix using the data from T265. The equations for this process are presented below and the transformations are visualized in Fig.\ref{transformations}. Subscription D stands for D435i sensor. Subscription T stands for T265 sensor.

\vspace{\baselineskip}
$ {}^{T}T_D$ - transformation between sensors, D435i w.r.t T265, always the same.

$ {}^{W}T_{T}$ - transformation of T265 sensor w.r.t world (actual data from T265 sensor)

$ {}^{W}T_{D} = {}^{W}T_{T} \cdot {}^{T}T_{D}$ - transformation of D435i sensor w.r.t world

${}^{D1}T_{D2}$ - transformation of KeyFrame2 w.r.t KeyFrame1 (transformation between poses of D435i - because we will use it to align PCs):

$ {}^{W}T_{D2} = {}^{W}T_{D1} \cdot {}^{D1}T_{D2}$

$ ({}^{W}T_{D1})^{-1} \cdot {}^{W}T_{D2} = {}^{D1}T_{D2}$

\begin{figure}[h]
\centering
\includegraphics[width=0.5\textwidth]{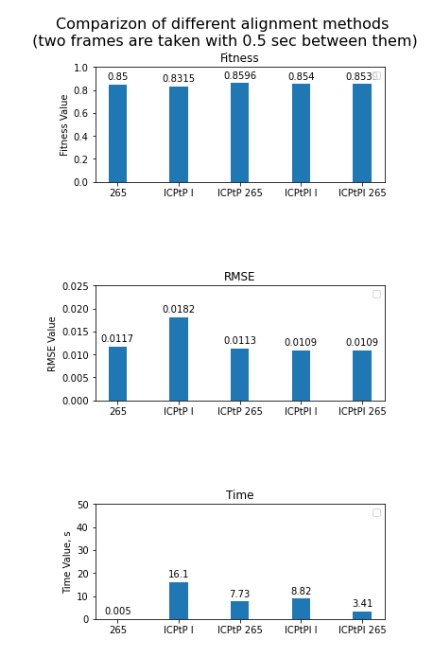}\label{true}
\caption{Comparison of (a) T265, (b) ICP p-t-point with initial identity, (c) ICP p-t-point with initial 265th transform, (d) ICP p-t-plane with initial identity, (e) ICP p-t-plane with initial 265th transform for PC alignment.}
\label{alighnment}
\end{figure}

To achieve a precise point cloud alignment in our project, we started with a comparison of different alignment approaches. For this project, two methods were chosen from the Iterative closest point (ICP) solution family \cite{ICP}: ICP Point-to-point and ICP Point-to-plane.  To evaluate their performance for the real-time mapping problem, we used RSME, fitness and time metrics. No downsampling and complex post-processing for pointclouds was performed in this first experiment. Both ICP methods were initialized at first with the identity matrix and then with a relative transformation from the 265s tracking camera as we wanted to evaluate not only their efficiency but also the improvement rate from our initial guess. The results of this experiment ( Fig.\ref{alighnment}) suggested that first: for the clear data the Point-to-plane approach over-performed Point to point in terms of both RMSE and time, and second: our initial guess from the 265th relative transformation doesn't always significantly affect RMSE (37,9\% in point but less then 1\% in plane method) but in both cases decrease the calculation time in about 50\% (the calculation of normals was also taken into account).

A key problem of real-time mapping algorithms is the constant compromising between computational accuracy and data processing time. For this reason, the post-processing influence on the two chosen approaches was additionally considered.

For this project the voxel grid downsampling was chosen that performs centroid estimation for every 4 pixels, as it represents an efficient real-time approach alongside the passthrough filtering and approximate voxel grid \cite{MorenoACS}. To evaluate other algorithms alongside the feature-preserving quadratic algorithm would actually be an interesting task, unfortunately, we didn't manage to finish their coding and had problems with PCL library installation to test their variations of passthrough, fast bilateral or statistical outlier removal that worked fine for group 3 task. 

\begin{figure}[h]
\centering
\includegraphics[width=0.4\textwidth]{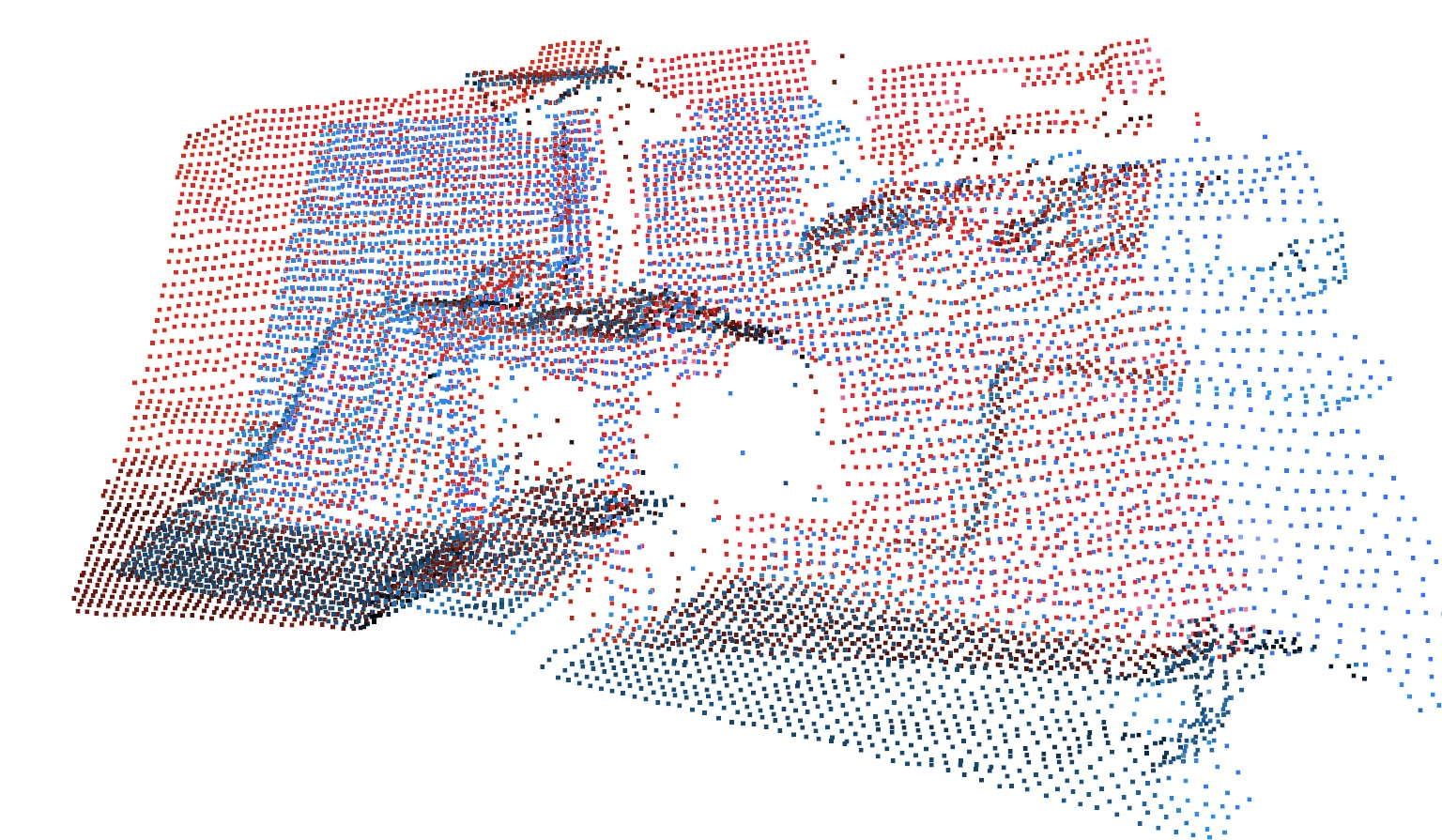}\label{true}
\caption{The result of voxel grid 4x4 pixel downsampling.}
\label{voxel}
\end{figure}

Alongside the downsampling approach, we tested a spatial edge-preserving filtering approach proposed in the research paper of Eduardo Gastal. Our initial hypotheses was that spatial filtering might work out partially similar to the passthrough by highlighting pointcloud's features and at the same time by smoothing process increase the efficiency of point cloud normal estimation , which took up most of the processing time for point-yo-plane.

\begin{figure}[h]
\centering
\includegraphics[width=0.4\textwidth]{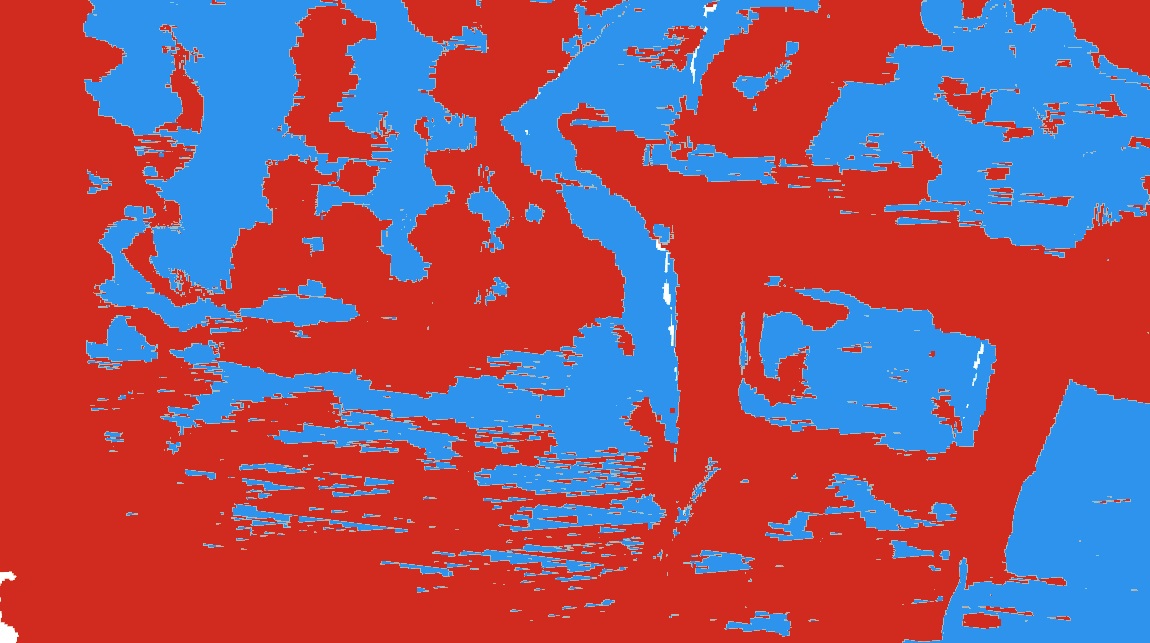}\label{true}
\caption{The result of spatial filter implementation.}
\label{spatial}
\end{figure}

\begin{figure}[h]
\centering
\includegraphics[width=0.5\textwidth]{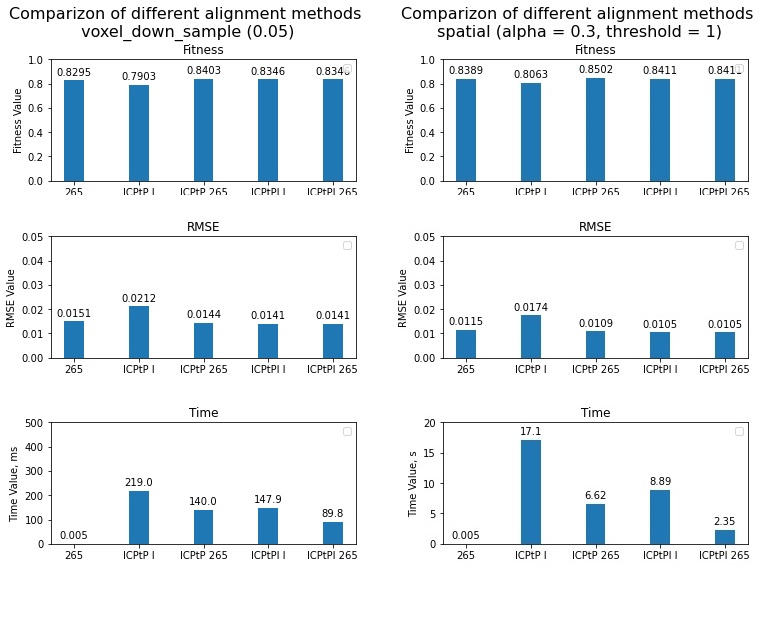}\label{true}
\caption{The ICP p-t-point and p-t-plane efficiency with the (a) voxel sampling, time in ms, (b) spatial filtering, time in s.}
\label{pc}
\end{figure}

The results of our experiment are presented in Fig.\ref{pc}  As can be seen on the graphs, after the voxel downsampling the processing time decreased by 50 times (90 ms in point-to-plane case) while RSME increased by 36\% , which still achieved a relatively low value for our task. At the same time, the spatial filtering didn’t considerably affect the overall alignment result even with a high smoothing factor which could be inflicted by either dense pointcloud formation or lower efficiency of tested algorithm in the 3D pointcloud case.

\section{Trajectory estimation}

During the experiment we have estimated trajectories using data from T265 and D435i sensors. Firstly, the trajectory was built depending on the transformation matrices obtained from T265 sensor. It was supposed that T265 sensor data is rather accurate. Nevertheless, it is prone to measurement errors. Therefore, it was decided to try to improve trajectory estimation by using observations of point clouds obtained by D435i sensor.

\begin{figure}[h]
\centering
\includegraphics[trim=100 0 0 30, clip, width=0.5\textwidth]{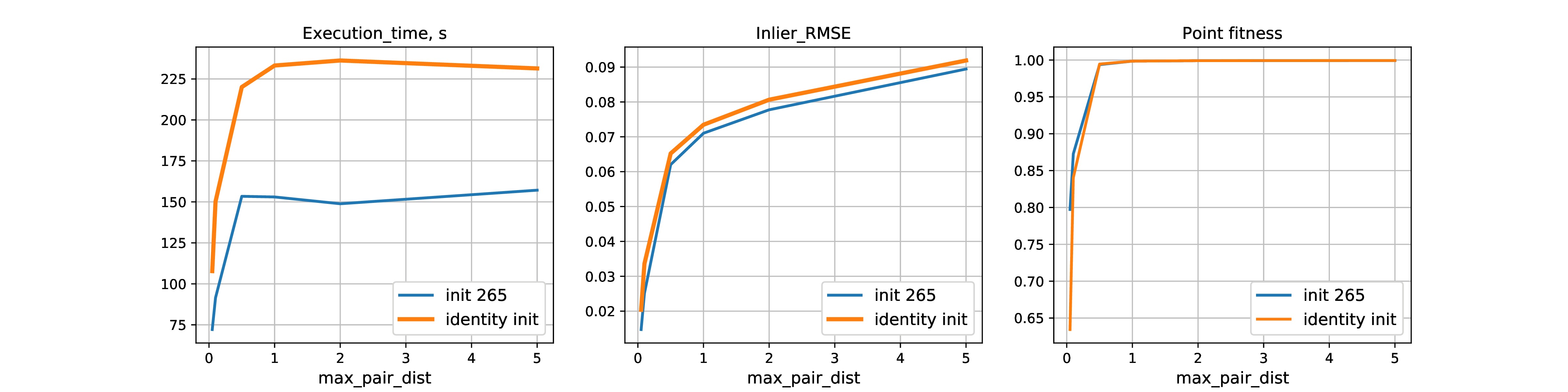}\label{true}
\caption{Comparison of different values of max pair dist.}
\label{max_pait_dist}
\end{figure}

To do this the transformation matrices obtained during the experiment with point clouds alignment were used. As it was described in the previous section, two approaches were considered. The first approach is when for ICP algorithm the transformation matrix is initialized by identity matrix. The second consisted of the initialization of the transformation matrix with the transformation matrix obtained from T265 sensor, i.e. the attempt to improve T265 transformation matrix taking into account observed point clouds. Also, for each initialization method, optimal maximum correspondence points-pair distance was picked up (Fig.\ref{max_pait_dist}). Depending on the received plots, maximum correspondence points-pair distance was set equal to 0.05 meters.

\begin{figure}[h]
\centering
\includegraphics[trim=180 90 90 90, clip, width=0.5\textwidth]{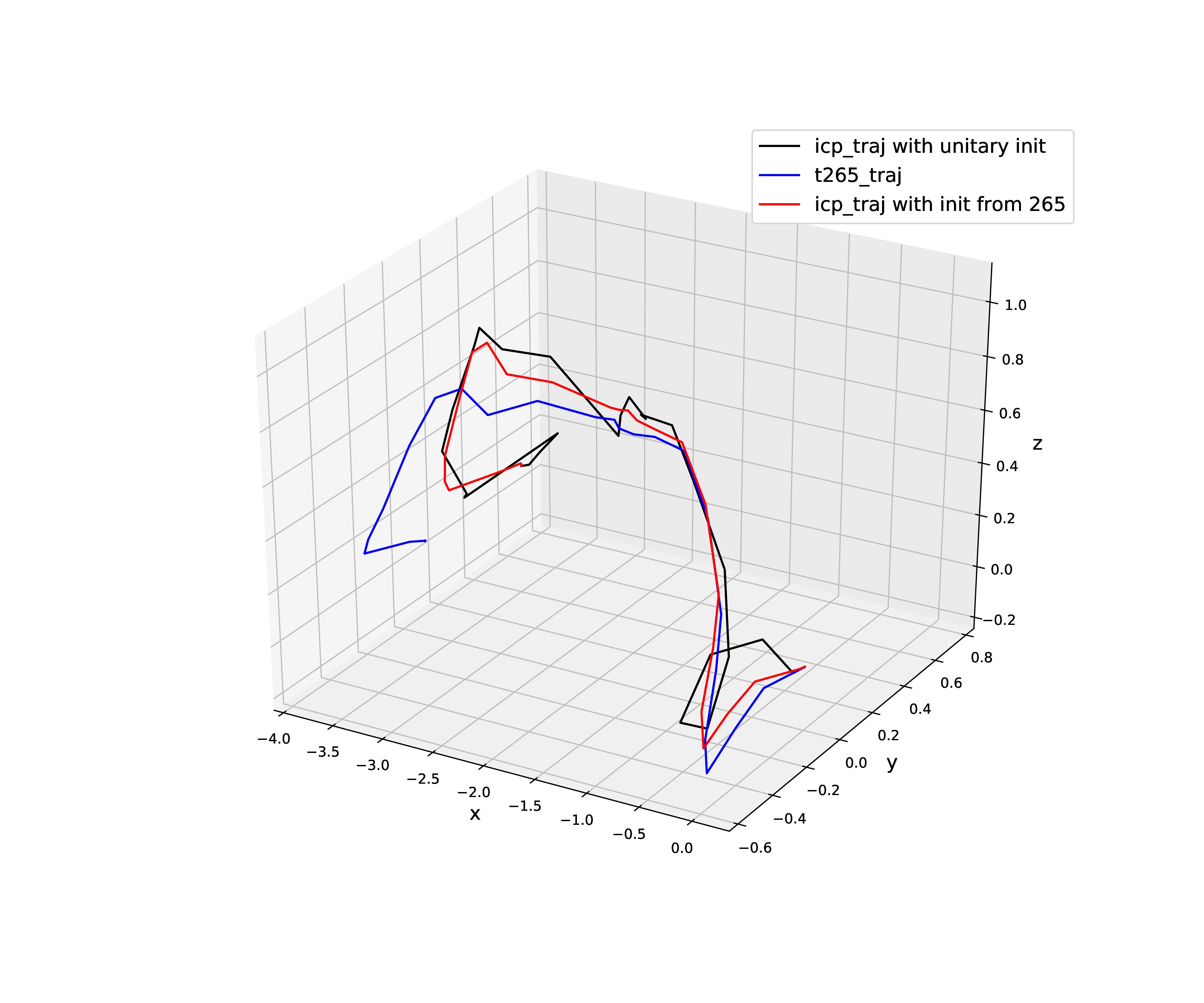}\label{true}
\caption{Comparison of trajectories obtained from pure T265 sensor, ICP with initial transformation matrix from T265 sensor and ICP with identity initial matrix.}
\label{trajectory}
\end{figure}

The built trajectories can be found on the Fig.\ref{trajectory}
As it is presented on the figure, the transformation matrices received from D435i sensor with identity matrix initialization give the worst results. Results for trajectories received from T265 sensor and D435i sensor with initialization by T265 matrices are comparable. Therefore, it was decided to use initialization by T265 sensor matrices for trajectory estimation by D435i sensor measurements.

\begin{figure}[h]
\centering
\includegraphics[trim=100 0 0 30, clip, width=0.5\textwidth]{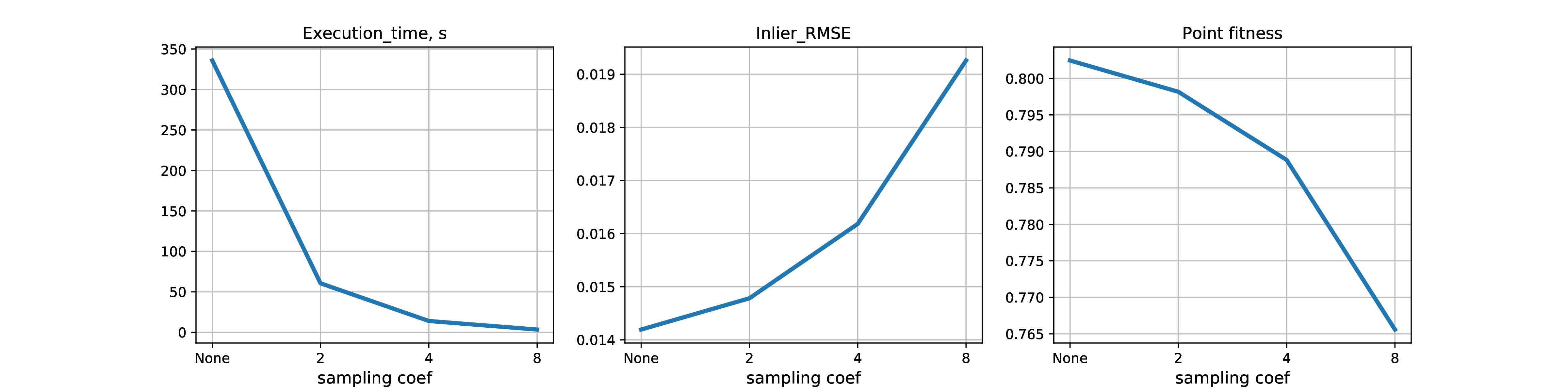}\label{true}
\caption{Comparison of ICP performance on different sampling coefficients of decimation filter.}
\label{sampl_ceof}
\end{figure}

\begin{figure}[H]
\centering
\includegraphics[trim=180 90 90 90, clip, width=0.5\textwidth]{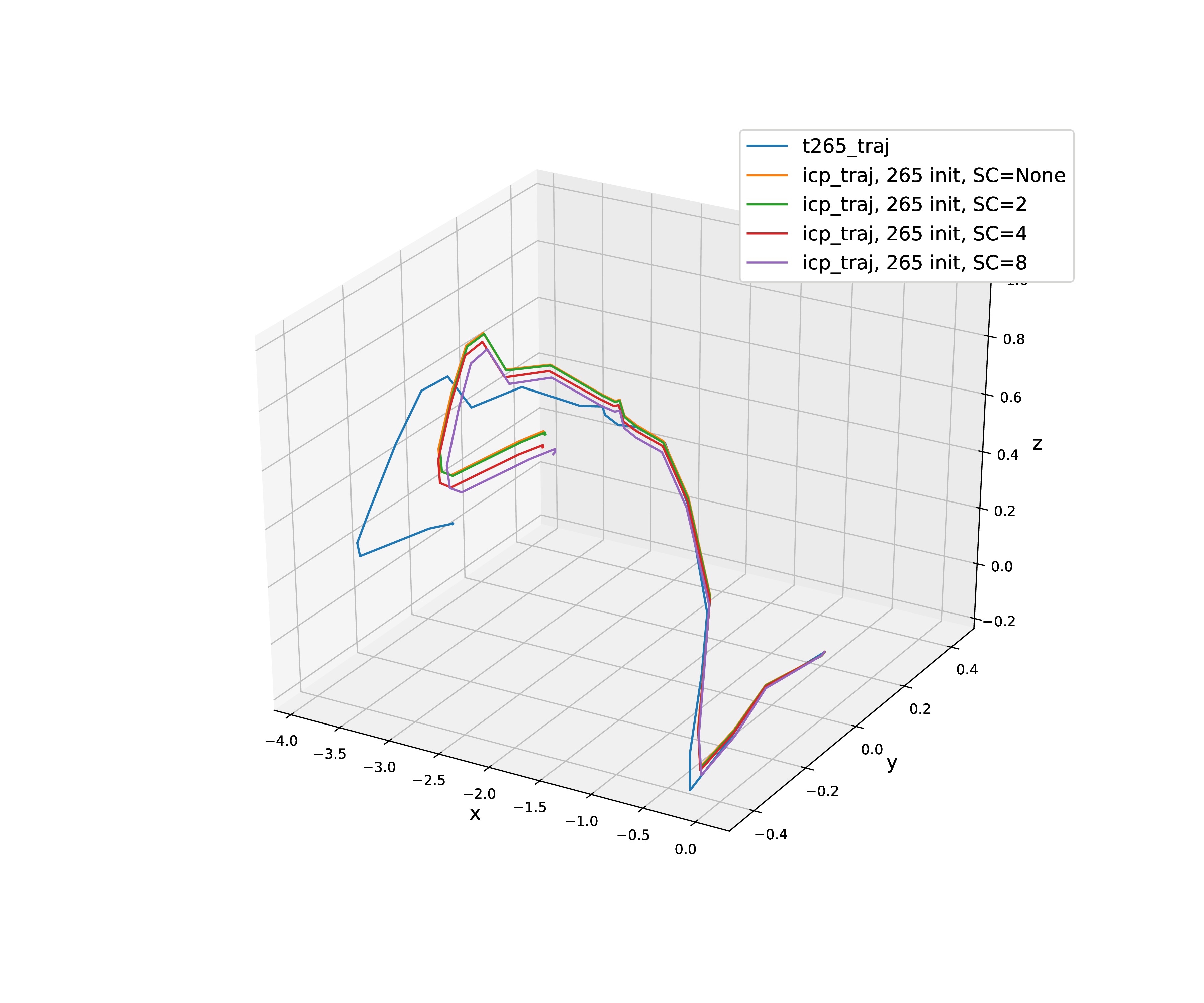}\label{true}
\caption{Comparison of the trajectories for different sampling coefficients of decimation filter.}
\label{traj_sampl_ceof}
\end{figure}

Nevertheless, the ICP algorithm has a drawback consisting in long time execution for big frames. In order to avoid this drawback, to the obtained from D435i sensor depth frames the decimation filter was applied (its implementation in \textit{pyrealsense2} library was used). To estimate optimal sampling coefficient experiments calculating time execution, RMSE and point fitness were carried out (Fig.\ref{sampl_ceof}). Sampling coefficient was chosen from the following values: $\{None, 2, 4, 8\}$, where $None$ means that decimation filter was not used. Corresponding trajectories are presented on the Fig.\ref{traj_sampl_ceof} Optimal trade-off between error of point clouds alignment and time execution give sampling coefficient equal to $4$.

\begin{figure}[h]
\centering
\includegraphics[trim=100 0 0 30, clip, width=0.54\textwidth]{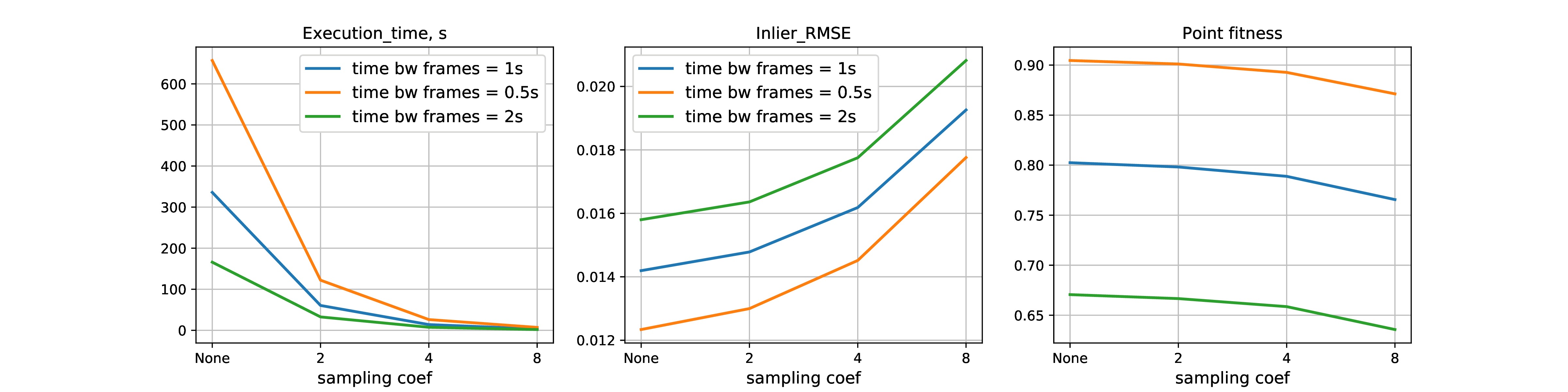}\label{true}
\caption{Comparison of ICP performance on different time interval between frames.}
\label{sampl_ceof_diff_tf}
\end{figure}

\begin{figure}[h]
\centering
\includegraphics[trim=180 90 90 90, clip, width=0.5\textwidth]{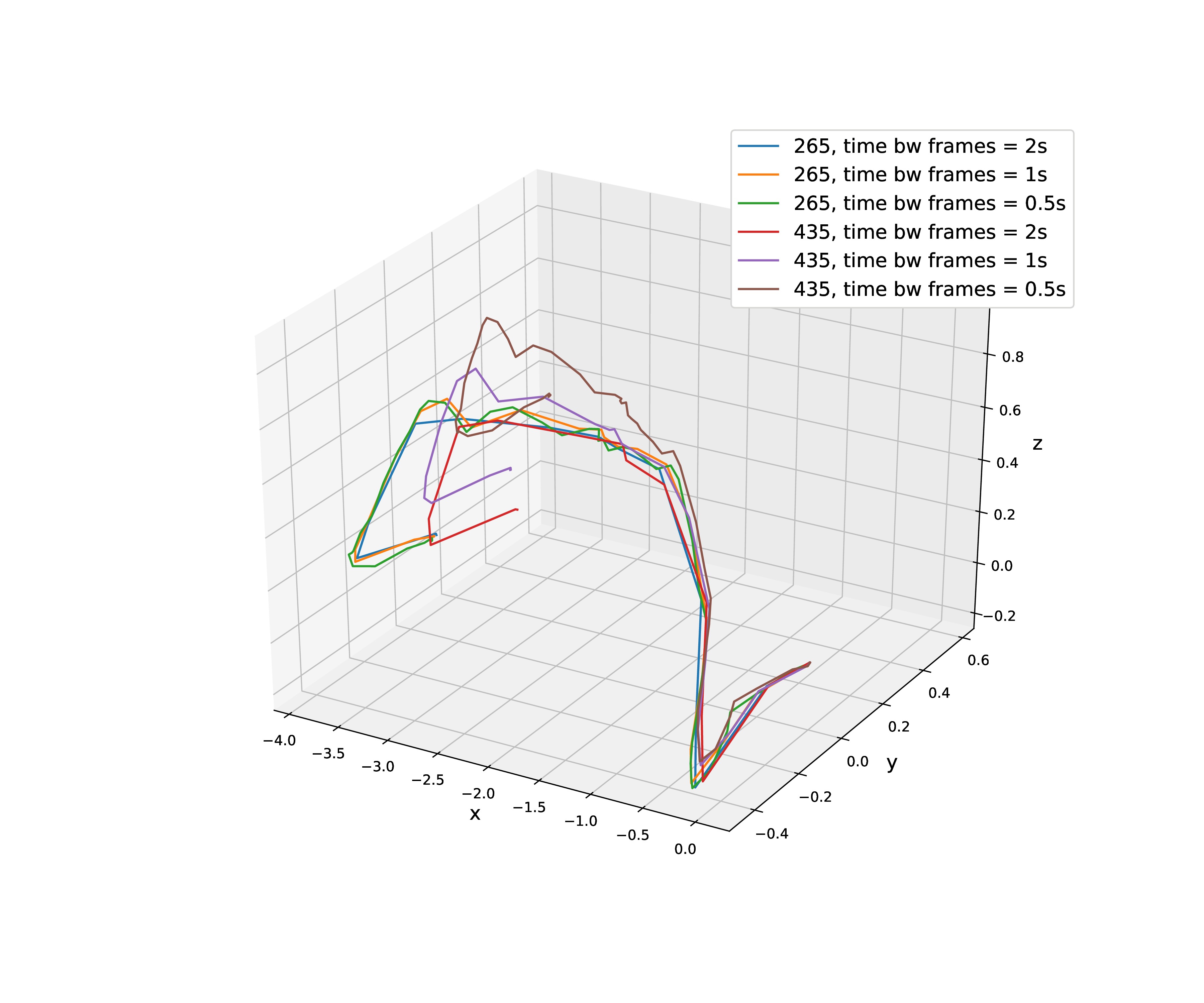}\label{true}
\caption{Comparison of estimated trajectories depending on the time interval between frames.}
\label{traj_diff_tf}
\end{figure}

The presented above results were obtained for frames of D435i sensor received every 1 second. To ensure that the change of interval between frames does not influence on the chosen optimal parameters, the same experiments were made for time intervals equal to $0.5$s and $2$s. Dependence on the sampling coefficient is presented on the Fig.\ref{sampl_ceof_diff_tf}. The corresponding trajectories for sampling coefficient equal to $4$ for each time interval between frames are shown on the Fig.\ref{traj_diff_tf}

Results obtained in this section show that for the trajectory estimation using data obtained from D435i sensor the optimal approach consist in using the ICP algorithm with the maximum correspondence points-pair distance equal to $0.05$ meters and transformation matrices initialization by transformation matrices received from T265 sensor. In order to decrease the execution time, the decimation filter with sampling coefficient equal to $4$ can be used. It does not deteriorate results significantly and gives acceptable trajectory estimations.

\begin{figure}[h]
\centering
\includegraphics[width=0.5\textwidth]{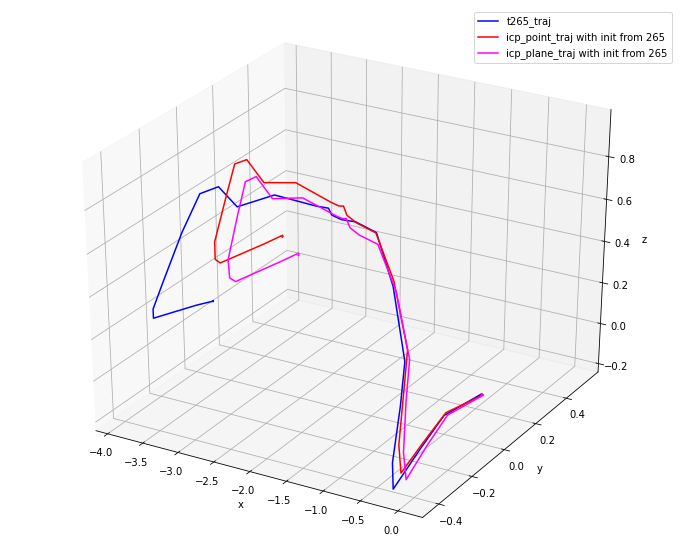}\label{true}
\caption{Comparison of estimated trajectories depending on the alignment method.}
\label{traj_ptpl}
\end{figure}

The additional comparison was performed to evaluate the results obtained for Point To Point and Point To Plane methods in terms of final trajectory estimation. The estimated results are presented in Fig.\ref{traj_ptpl}

As it can be found from the plots, both ICP methods performed a similar trajectories it terms of relative translations and rotations. However, Point To Point method performed less relative deviation in the x-axis from the trajectory obtained by tracking sensor. Overall, without ground truth it yet not possible to make a final suggestion on which alignment algorithm was more precise.

\section{Mapping}

\begin{figure}[h]
\centering
\includegraphics[clip, width=0.5\textwidth]{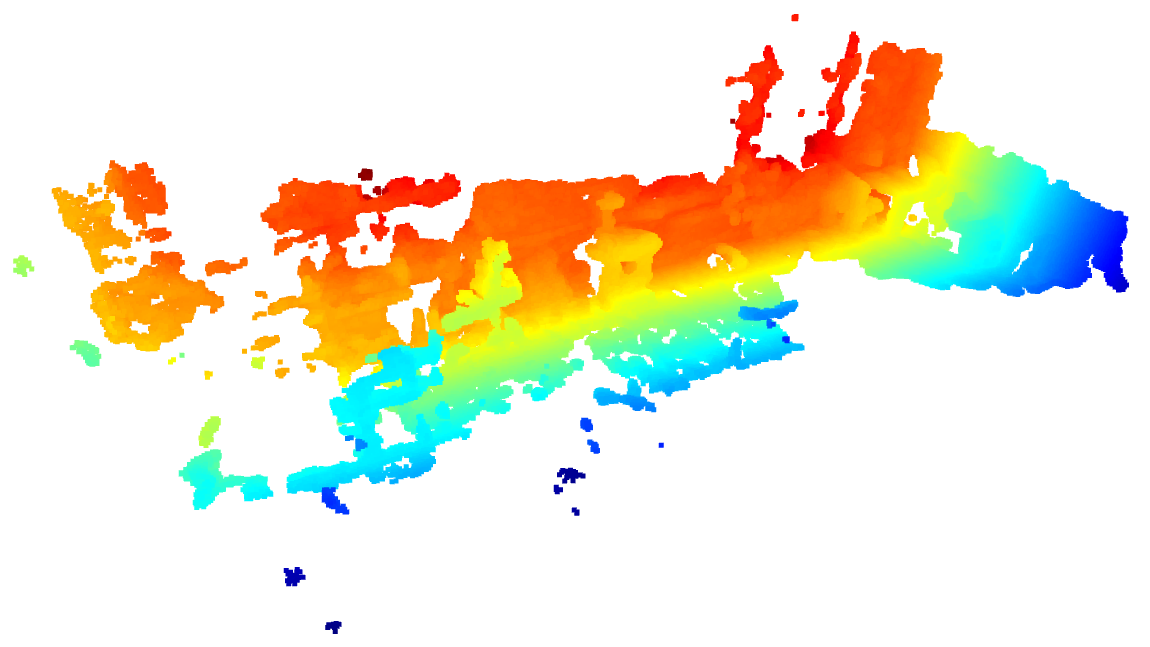}\label{true}
\caption{Open3D. Resultant point cloud.}
\label{res_point_cloud}
\end{figure}

For our project we recorded some data sets by both D435i and T265 sensors working simultaneously. So, some area was observed by the sensors. One of our tasks was to reconstruct the whole map of the observed area using obtained data sets.
First of all, in order to achieve the goal, we needed a global set of point clouds to be merged and built in one initial coordinate system. To do that we combined corresponding frames iteratively using already computated transformation matrices ${}^{D1}T_{D2}$ between KeyFrames mentioned earlier. All operations were done through \textit{Open3D} library was used. The result of the combinations is provided on Fig.\ref{res_point_cloud}. For the better visualization downsampling was conducted. A shape of the resultant point cloud fully corresponds to the real world which allows to conclude that all transformations were done correctly.

\begin{figure}[H]
\centering
\includegraphics[clip, width=0.5\textwidth]{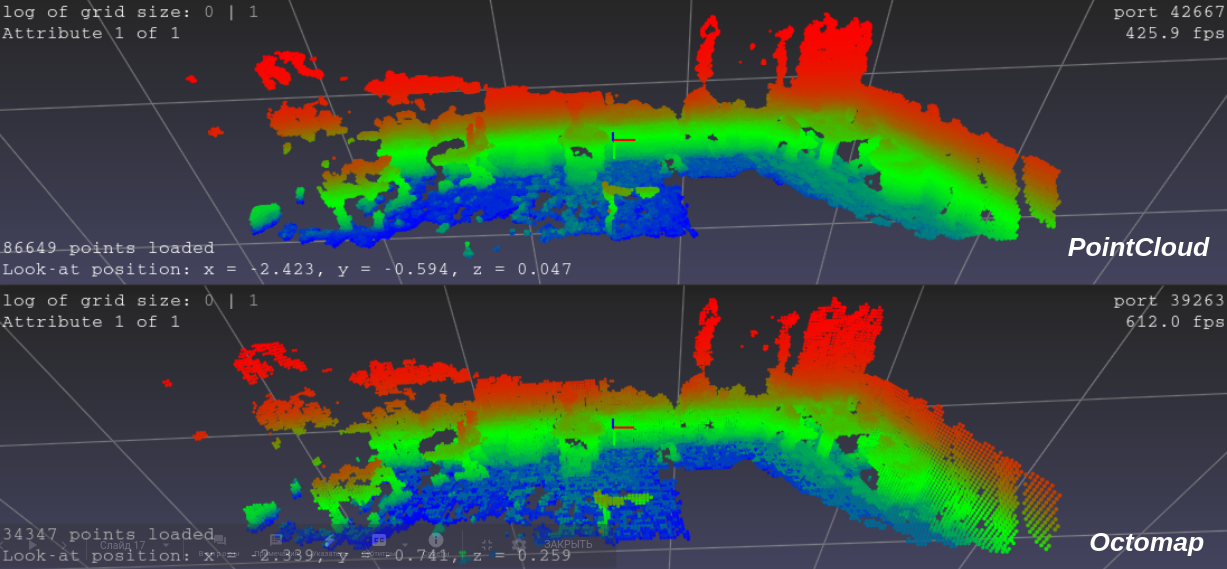}\label{true}
\caption{PPTK. Compare of resultant point cloud with the map.}
\label{pcl_vs_oct}
\end{figure}

The next step was a reconstruction of the map itself using \textit{Octomap} module. We decided to avoid usage of \textit{Octomap} with \textit{ROS (Robotic Operating System)} in order to make the representation more adaptive to different operating systems. \textit{Octomap} was chosen because of its clarity and stable work.
For the map visualization, \textit{PPTK} module was chosen first. On the Fig.\ref{pcl_vs_oct} a compare of the colored resultant point cloud with a reconstructed map is provided.

\begin{figure}[H]
\centering
\includegraphics[clip, width=0.5\textwidth]{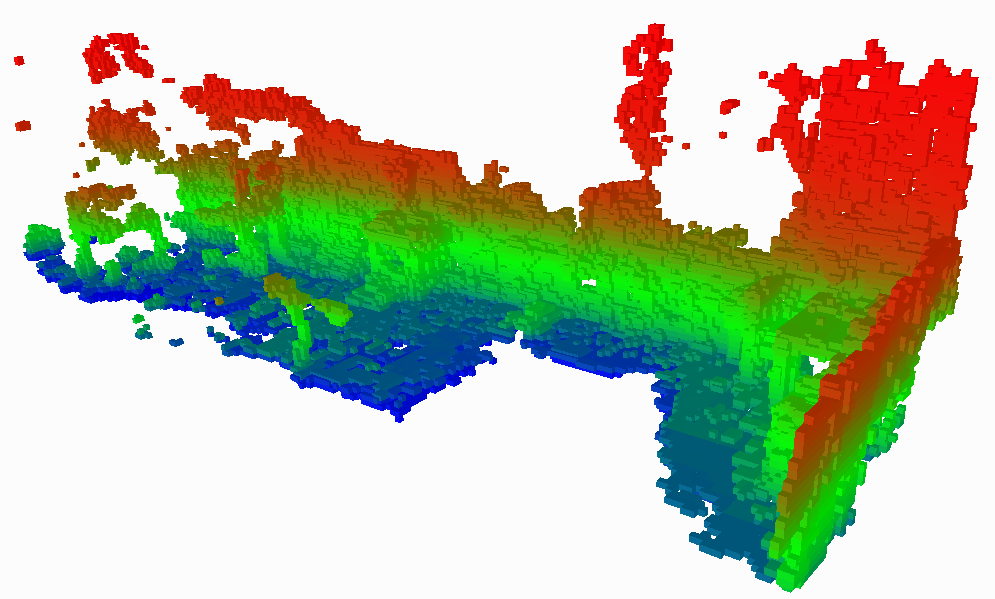}\label{true}
\caption{Pyglet. Traditional colored resultant map with voxel representation.}
\label{map_cubes}
\end{figure}

As it is shown on the figure above, \textit{Octomap} provides a good calculation of occupied points which are used for the map representation. However, \textit{PPTK} module doesn't provide a traditional representation of 3D map with voxels.
So, the traditional colored representation was realized with \textit{pyglet} module, Fig.\ref{map_cubes} The colors from blue to red shows a direction from the floor to the ceiling.

\section{Future work}

For the future work we propose to build a filter to solve a localization problem in a better way. Tracking data from T265 camera can be treated as a source of optometry with low level of uncertainty. While the RGBD data from D435i camera can be treated as an observations with high uncertainty level. EKF or Particle Filter can be applied here.

In order to be able to navigate a robot in real time we need to update the map fast enough \cite{Guizilini_2018}. That is extremely important for robots with fast dynamics, such as drones. In order to meet this requirements, we propose to implement mapping using acceleration with CUDA GPU. In the work of Pan et. al. \cite{pan2019gpu} presented approach that allows to make GPU acceleration and multi-threaded parallelism for map calculation. In their approach they used data from LiDAR sensor, so our final solution may have a different approach.

Accelerate the map exploration of the robot by using auto-encoders/GANs that complete map and predict its optimal trajectory~\cite{learned_map}.








\bibliographystyle{ieeetr}
\bibliography{literature.bib}

\end{document}